\documentclass[letterpaper]{article} 
\usepackage{aaai2027}  
\usepackage[hyphens]{url}  
\usepackage{graphicx} 
\usepackage{natbib}  
\usepackage{caption} 
\usepackage{algorithm}
\usepackage{algorithmic}

\usepackage{newfloat}
\usepackage{listings}
\DeclareCaptionStyle{ruled}{labelfont=normalfont,labelsep=colon,strut=off} 
\floatstyle{ruled}
\newfloat{listing}{tb}{lst}{}
\floatname{listing}{Listing}

\usepackage{booktabs}

\usepackage{multirow}
\usepackage{amsmath}
\usepackage{amssymb}
\usepackage[table]{xcolor}
\nocopyright 

\title{AdvTiles: Physical Adversarial Camouflage Clothing against Person Detectors via Learnable Tiles}

\author{
    Jinlei Wang\textsuperscript{\rm 1,\rm 2}\equalcontrib,
    Jiahuan Long\textsuperscript{\rm 3}\equalcontrib,
    Mingkai Sun\textsuperscript{\rm 2},
    Yafei Guo\textsuperscript{\rm 2},
    Yuanhao Huang\textsuperscript{\rm 5},
    Ming Wang\textsuperscript{\rm 2,\rm 4},\\
    Junqi Wu\textsuperscript{\rm 2,\rm 4},
    Jiacheng Hou\textsuperscript{\rm 2},
    Hongbo Chen\textsuperscript{\rm 1},
    Xingxing Wei\textsuperscript{\rm 5},
    Tingsong Jiang\textsuperscript{\rm 2}\corresponding,
    and Wen Yao\textsuperscript{\rm 2}\corresponding
}

\affiliations{
    \textsuperscript{\rm 1} Sun Yat-sen University
    ~\textsuperscript{\rm 2}Chinese Academy of Military Science\\
    \textsuperscript{\rm 3}Shenzhen University
    ~\textsuperscript{\rm 4}Shanghai Jiao Tong University
    ~\textsuperscript{\rm 5}Beihang University
}

\begin{document}

\maketitle

\begin{abstract}
Physical adversarial attacks against person detectors have evolved from localized patches to full-body textures. However, achieving both visual naturalness and strong attack effectiveness remains challenging. Existing natural-looking methods typically optimize camouflage textures as a whole, limiting the flexibility to refine local adversarial patterns and their spatial arrangement. To address this issue, we  propose AdvTiles, a physical adversarial camouflage framework built from learnable tiles, enabling strong attack performance while preserving a natural camouflage appearance. Specifically, we use a  Straight-through (ST) Gumbel-Softmax estimator for differentiable tile selection, enabling joint optimization of tile patterns and spatial layouts. This design provides fine-grained control over adversarial texture generation. To improve robustness in diverse physical conditions, we further optimize the camouflage through differentiable 3D Gaussian Splatting rendering with variations in viewpoints, scales, illuminations and backgrounds. Extensive experiments across multiple detectors demonstrate that AdvTiles achieves an average ASR of 86.2\%, outperforming existing state-of-the-art attack methods. We further fabricate the optimized camouflage into wearable adversarial clothing, validating its effectiveness in real-world scenarios across diverse distances, angles and backgrounds.
\end{abstract}

\section{Introduction}
Deep neural networks have substantially advanced person detection and are now widely deployed in intelligent surveillance and other safety-critical vision systems~\cite{lei2025yolov13realtimeobjectdetection,huang2025deim, long2025robust, zhao2024detrs, yao2026camotion}. Despite their strong performance, deep-learning based person detectors remain vulnerable to adversarial examples, as subtle perturbations can induce false positives or missed detections. In real-world deployments, such vulnerabilities in person detectors may directly undermine the reliability of intelligent surveillance and security systems~\cite{wang2024adversarialexamplesphysicalworld, chakraborty2021survey}.

\begin{figure}[t]
    \centering
    \includegraphics[width=0.95\linewidth]{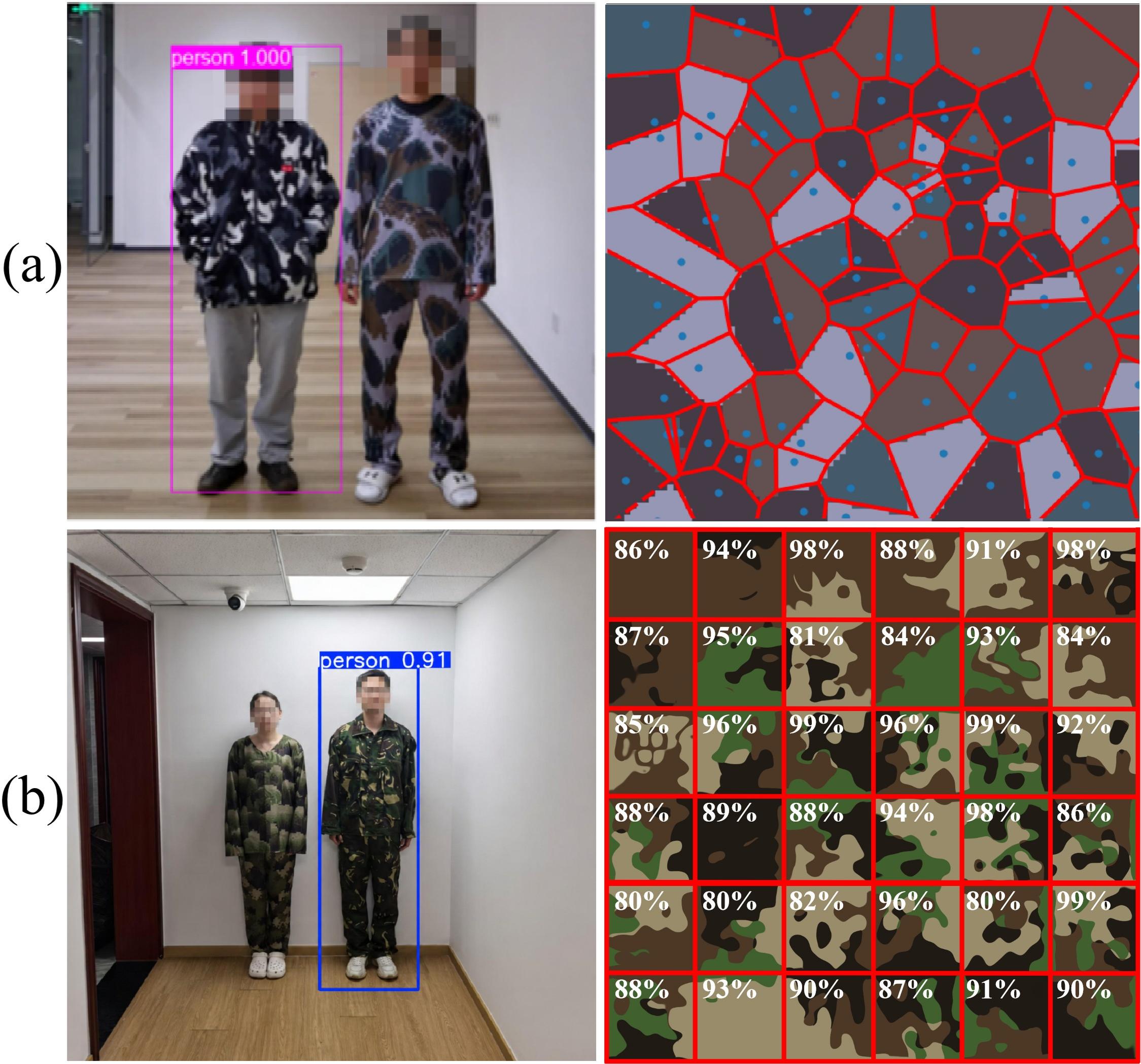}
    \vspace{-0.08in}
    \caption{Comparison of representative adversarial camouflage methods.
    (a) Previous method~\cite{hu2023physically} represents adversarial camouflage using Voronoi diagrams parameterized by learnable control points.
    (b) Our method represents adversarial camouflage as an arrangement of tiles, where tile selection probabilities and spatial layouts are jointly optimized to enable finer-grained adversarial texture. 
    }
    \label{fig-cover}
\end{figure}

Recently, physical adversarial attacks have attracted increasing attention for evaluating the real-world robustness of person detectors. Existing attacks against person detectors can be broadly categorized into two types: patch-based attacks~\cite{thys2019fooling,cheng2024full,huang2026advreal} and texture-based adversarial attacks~\cite{hu2023physically,hou2025fab}. Patch-based adversarial attacks typically optimize localized adversarial patterns attached to clothing. For instance, Thys et al.~\cite{thys2019fooling} designed a printable patch to hide persons from object detectors. However, the limited spatial coverage of adversarial patch makes it difficult to maintain effective attacks across varying viewpoints and body poses.

To improve attack robustness across diverse viewpoints, texture-based adversarial methods extend adversarial patterns over larger clothing surfaces~\cite{hu2023physically}. A pioneering work in this direction is AdvTexture~\cite{hu2022adversarial}, which generates expandable adversarial textures that can cover clothing of different shapes and remain effective across multiple viewpoints. Building on this line of research, subsequent studies have further explored natural-looking adversarial texture to reduce the visual conspicuousness of adversarial clothing. For example, Hu et al.~\cite{hu2023physically} proposed natural camouflage textures using Voronoi diagrams parameterized by learnable control points. However, such a point-control-based method typically optimizes camouflage textures as a whole, limiting the flexibility to refine local adversarial patterns and their spatial arrangement. Consequently, \textbf{achieving both visual naturalness and strong attack effectiveness in adversarial camouflage remains challenging.}

In this paper,  we propose AdvTiles, a physical adversarial camouflage framework built from learnable tiles, enabling strong attack performance while preserving a natural camouflage appearance. Instead of directly optimizing a whole adversarial texture, AdvTiles represents the texture as a layout of tiles. It leverages a Straight-through
Gumbel-Softmax estimator for differentiable tile selection, enabling joint optimization of tile patterns and spatial layouts. This design provides finer-grained control over adversarial texture generation. As shown in Figure~\ref{fig-cover}, the representative adversarial camouflage method AdvCaT generates the adversarial texture by optimizing the control points. In contrast, our method enables finer-grained textures by jointly optimizing individual tile patterns and their spatial arrangement.   
Moreover, to improve robustness in diverse physical conditions, we further optimize the camouflage through differentiable 3D Gaussian Splatting rendering with variations
in viewpoints, scales, illuminations and backgrounds. Our main contributions are summarized as follows:

\begin{itemize}

\item We propose a physical adversarial camouflage framework against person detectors. It represents adversarial textures as arrangements of learnable tiles to achieve both strong attack effectiveness and a natural-looking appearance.

\item We introduce 
differentiable tile optimization via ST Gumbel-Softmax for jointly learning tile patterns and spatial layouts. We further leverage 3D Gaussian Splatting to enhance the camouflage  robustness under diverse physical conditions. 

\item Digital experiments across multiple detectors demonstrate that our method consistently outperforms existing SOTA methods in attacking person detectors. Physical experiments further validate its real-world effectiveness under diverse distances, angles and backgrounds. 

\end{itemize}

\section{Related Work}
\noindent\textbf{Patch-Based Physical Adversarial Attacks.}
Existing digital attacks ~\cite{moosavi2016deepfool,chen2018ead,xie2019improving,li2020learning, wei2024revisiting, long2025robust, wang2025transferable, zhou2025fooling, huang2026advreal} demonstrated that subtle digital perturbations can severely mislead deep neural networks. 
To extend adversarial attacks to the physical world, Brown et al.~\cite{brown2017adversarial} introduced universal adversarial patches, which can be printed and placed in real-world scenes to fool image classifiers. Building on this work, Thys et al.~\cite{thys2019fooling} further extended adversarial patches from image classification to person detection by suppressing the detection confidence. 
Subsequent studies further improved adversarial patches in terms of visual naturalness, transferability, and physical robustness~\cite{hu2021naturalistic, huang2023t, guesmi2024dap, long2024papmot, chen2025enhancing, long2025cdupatch, hu2025dynamicpae, long2026thermally, yan2026diff}. Hu et al.~\cite{hu2021naturalistic} generated natural-looking patches using the latent space of a generative model. Huang et al.~\cite{huang2023t} introduced self-ensemble optimization to improve black-box transferability. Guesmi et al.~\cite{guesmi2024dap} employed dynamic patches and crease transformations to accommodate clothing deformation. More recently, Yan et al.~\cite{yan2026diff} leveraged class-optimized diffusion models to generate adversarial patches with improved visual naturalness and attack effectiveness. 
However, patch-based attacks apply adversarial patterns to only a small part of the human body. As a result, their effectiveness can easily degrade under viewpoint changes.

\begin{figure*}[t]
    \centering
    \includegraphics[width=0.95\linewidth]{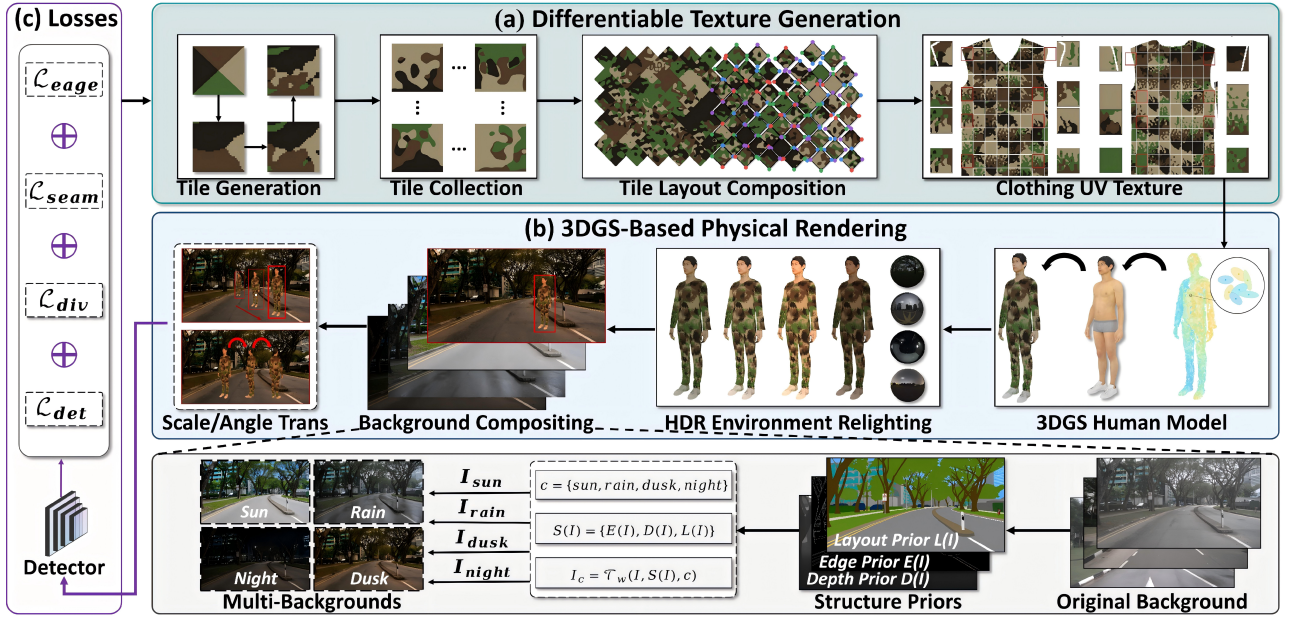}
    \vspace{-0.05in}
    \caption{
    Overview of the AdvTiles framework.
    (a) Differentiable camouflage generation. Learnable tile patterns are optimized and composited into UV texture for the t-shirt and trouser. (b) 3DGS-based Physical Rendering. The generated UV textures are mapped onto the clothing regions of a 3D Gaussian Splatting (3DGS) human model,
    followed by HDR environment relighting, background compositing, and viewpoint and scale transformations. (c) Adversarial Losses. The rendered images are fed into the detector, and the adversarial losses are backpropagated to jointly optimize the tile patterns and spatial layout.
    }
    \label{fig-overview}
\end{figure*}

\noindent\textbf{Texture-Based Physical Adversarial Attacks.} 
To improve multi-view attack robustness, texture-based methods extend adversarial patches over larger clothing surfaces~\cite{9009841, xu2020adversarial, raffiee2021garmentgan, sun2023differential, zhang2025single, zhu2026physical,huang2026advserial, zhu2026rfa}.
Hu et al.~\cite{hu2022adversarial} proposed AdvTexture, which repeatedly expands adversarial textures to cover clothing of different shapes and improve attack stability across multiple viewpoints. 
Beyond improving attack robustness, recent works have focused on generating more natural-looking textures to reduce the visual conspicuousness.
Lu et al.~\cite{10.1145/3658664.3659630} proposed AdvOcl, which leverages the learned image manifold of the diffusion model to generate patterns for daily clothes.
Sun et al.~\cite{sun2023differential} proposed DE-DAC, which generates scene-adaptive camouflage textures that visually blend with the surrounding environment. However, both methods primarily focus on digital evaluation and are difficult to deploy in the physical world.
In contrast, Hu et al.~\cite{hu2023physically} proposed AdvCaT, which parameterizes adversarial textures using Voronoi patterns with learnable control points, producing natural-looking camouflage that can be physically fabricated and deployed.
However, in Voronoi-based methods, local regions are implicitly determined by a limited set of control points, making it difficult to independently refine fine-grained texture details within each region.  

In this paper, we propose AdvTiles, a physical adversarial camouflage framework based on learnable tiles. Instead of directly optimizing the entire adversarial texture, we represent it as an arrangement of learnable tiles, enabling strong attack performance while preserving a natural camouflage appearance.

\section{Methodology}
In this section, we present the AdvTiles framework for generating a natural-looking adversarial camouflage texture that can be mapped onto clothing to evade person detectors. As shown in Figure~\ref{fig-overview}, it consists of three key modules: (a) Differentiable Texture Generation, (b) 3DGS-based Physical Rendering, and (c) Adversarial losses, which are detailed below.

\subsection{Differentiable texture Generation}

Existing adversarial camouflage methods~\cite{hu2023physically, huang2026advreal} often optimizes the clothing texture as a whole, making their attack effectiveness dependent on the integrity of the global pattern and thus sensitive to physical-world variations, such as clothing deformation, partial occlusion and scale changes.

To overcome this limitation, we represent the entire clothing texture as an arrangement of learnable tiles. Each tile has its own learnable parameters, helping preserve local adversarial cues under physical-world variations. Figure~\ref{fig-overview}(a) illustrates the differentiable texture generation process. It first optimize the color regions and boundaries of multiple learnable tiles to form a tile collection, from which differentiable tile selection assigns one tile to each grid cell, thereby composing a complete tile layout. The resulting layout is used to construct the final clothing UV texture.  The details of tile generation and tile layout composition are described below.

\begin{figure}[t]
    \centering
    \includegraphics[width=0.9\linewidth]{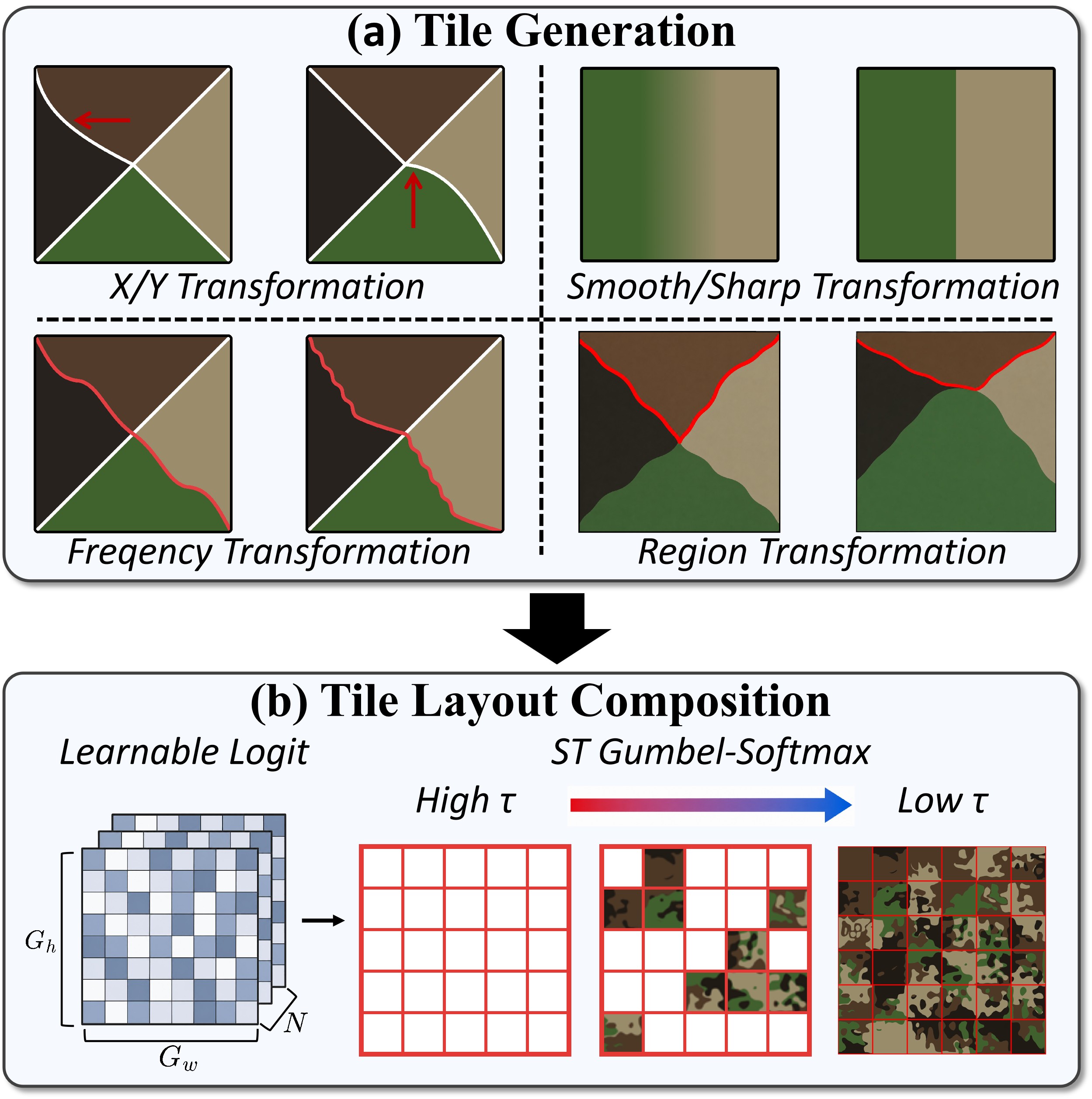}
    \vspace{-0.08in}
    \caption{Differentiable camouflage space.
    (a) Four learnable transformations for tile generation.
    (b) Layout composition via learnable tile-assignment logits and straight-through Gumbel-Softmax.
    }
    \label{fig-dcs}
\end{figure}

\noindent\textbf{Tile Generation.} 
We define the camouflage color set $\mathcal{C}$ as:
\begin{equation}
\mathcal{C}
=
\left\{
c_j=(r_j,g_j,b_j)\mid j=1,\ldots,N_C
\right\},
\end{equation}
where $N_C$ represents the number of candidate colors and 
$c_j$ denotes the RGB triplet of the $j$-th color.
Let $\mathcal{B}=\{B_i\}_{i=1}^{N}$ denote a collection of $N$
learnable tiles. Each tile $B_i$ is associated with four edge labels:
\begin{equation}
k_i=(k_i^l,k_i^r,k_i^t,k_i^b),
\end{equation}
where $k_i^l$, $k_i^r$, $k_i^t$, and $k_i^b$ denote the symbolic keys of the left, right, top, and bottom boundaries of tile $B_i$, respectively.
To ensure seamless composition, adjacent tiles must have matching
keys along their shared edges.
Specifically, if tile $B_j$ is placed to the right of $B_i$,
horizontal compatibility requires $k_i^r=k_j^l$.
Similarly, if $B_j$ is placed below $B_i$, vertical compatibility
requires $k_i^b=k_j^t$.
For a normalized coordinate $p=(p_x,p_y)\in[0,1]^2$ within tile
$B_i$, the RGB color at $p$ is defined as
\begin{equation}
B_i(p)=G\left(p;k_i,\theta_i,\mathcal{C}\right),
\end{equation}
where $G$ denotes the differentiable tile generator and
$\theta_i=(\alpha_i,s_i,f_i,\beta_i)$ denotes the learnable
transformation parameters of tile $B_i$. 
As shown in Figure~\ref{fig-dcs}(a), the parameters $\alpha_i$, $s_i$, $f_i$, and $\beta_i$ control
X/Y translation, boundary sharpness, contour frequency, and
region size, respectively.
$\alpha_i$ shifts the color regions along the horizontal and vertical directions.
$s_i$ controls the transition
between adjacent regions. $f_i$ regulates
contour oscillations and thus determines boundary complexity. $\beta_i$ adjusts the relative area occupied by each color region.

\noindent\textbf{Tile Layout Composition.}
As illustrated in Figure~\ref{fig-dcs}(b), we compose the generated
tiles into a UV texture by learning a tile-assignment logit tensor
$Z\in\mathbb{R}^{N\times G_h\times G_w}$, where $Z_{i,y,x}$ denotes
the score of assigning tile $i$ to grid cell $(y,x)$, and
$G_h\times G_w$ is the tile-grid size.
Since selecting one tile per cell is discrete, we use the
Straight-through (ST) Gumbel--Softmax estimator
~\cite{jang2017categorical}:
\begin{equation}
\begin{aligned}
P_{:,y,x}
&=
\operatorname{softmax}
\left(
\frac{Z_{:,y,x}+g_{:,y,x}}{\tau}
\right),\\
H_{:,y,x}
&=
\operatorname{onehot}
\left(
\operatorname*{arg\,max}_{i}P_{i,y,x}
\right),\\
\widehat{H}
&=
\operatorname{sg}(H-P)+P.
\end{aligned}
\end{equation}

Here, $P$ is the relaxed assignment, $H$ is its hard one-hot
counterpart, $g$ is standard Gumbel noise, $\tau$ is the temperature,
and $\operatorname{sg}(\cdot)$ denotes stop-gradient.
Thus, $\widehat{H}$ uses the hard assignment in the forward pass while
propagating gradients through $P$.
We gradually decrease the temperature value $\tau$ during optimization to make $P$ approach
a one-hot distribution.

The selected tiles are arranged to form the UV texture $T$:
\begin{equation}
T_{:,u,v}
=
\sum_{i=1}^{N}
\widehat{H}_{i,y,x}B_i(p_{u,v}),
\quad
(u,v)\in\Omega_{y,x},
\end{equation}
where $\Omega_{y,x}$ denotes the pixel region of grid cell $(y,x)$,
$p_{u,v}$ is the normalized local coordinate within the tile, and
$T_{:,u,v}$ is the RGB vector at pixel $(u,v)$.

\subsection{3DGS-based Physical Rendering}
As described above, tile layout composition produces 2D UV textures for the clothing.
However, direct optimization in the 2D texture space does not account for 3D projection and viewpoint changes during physical deployment.
As shown in Figure~\ref{fig-overview}(b), we map these textures onto the predefined clothing regions of a 3D Gaussian Splatting (3DGS)~\cite{10.1145/3592433} human model using UV coordinates.
The textured model is then rendered under sampled viewpoints and scales,
relit using HDR environment relighting, and composited with diverse backgrounds
to simulate physical-world observations.

\noindent\textbf{3DGS Human Model.}
The human model is represented by a set of anisotropic Gaussian
primitives $\mathcal{G}$.
We freeze its geometry, covariance, opacity, and non-clothing
appearance, and optimize only the tile parameters that define the
clothing texture $T$.
The texture is mapped onto the clothing Gaussians using precomputed
UV coordinates, yielding the textured model $\mathcal{G}(T)$. Given a sampled camera configuration $\theta_c$, including the viewpoint
and observation scale, the differentiable 3D Gaussian Splatting renderer (3DGR) produces the
human image $\widetilde{I}_r$ and its foreground alpha mask $M_r$:
\begin{equation}
(\widetilde{I}_r,M_r)
=
R_{\mathrm{3DGS}}
\left(
\mathcal{G}(T),\theta_c
\right),
\end{equation}
where $R_{\mathrm{3DGS}}$ denotes the 3DGS renderer with
view-independent colors. We then relight the rendered human using a sampled high-dynamic-range
(HDR) environment map $\ell$:
$I_r
=
A_{\mathrm{hdr}}
\left(
\widetilde{I}_r,\ell
\right),
$ where $A_{\mathrm{hdr}}$ denotes the illumination modulation operator.
Since both operations are differentiable, detector gradients can
propagate through $I_r$ to the tile appearance parameters and
tile-assignment logits.

\noindent\textbf{Background Relighting and Compositing.} 
To simulate diverse environmental conditions, we use IC-Light
~\cite{zhang2025scaling} to generate multiple weather and illumination variants
of each background image.
Given an original background
$I_{o}$, we extract a set of structural priors:
$S(I_{o})
=
\left\{
E(I_{o}),
D(I_{o}),
L(I_{o})
\right\},$
where $E(I_{o})$, $D(I_{o})$, and
$L(I_{o})$ denote the edge, monocular depth, and scene-layout
priors extracted from $I_o$, respectively. For each condition $c \in \mathcal{C}_{\mathrm{env}}$, where
$\mathcal{C}_{\mathrm{env}}
=
\{\mathrm{sun}, \mathrm{rain}, \mathrm{dusk}, \mathrm{night}\}$,
the transformation function $\mathcal{T}_w$ generates a
corresponding background:
$I_{c}
=
T_{w}
\left(
I_{o},
S(I_{o}),
c
\right).
$ 
The resulting background pool is defined as
$\mathcal{P}_{B}
=
\left\{
I_{c}
\mid
I_{o}\in\mathcal{D}_{B},
c\in\mathcal{C}_{env}
\right\},$
where $\mathcal{D}_{B}$ denotes the collection of original
backgrounds. During optimization, a background
$B\sim\operatorname{Uniform}(\mathcal{P}_{B})$ is
randomly sampled and composited with the rendered human:
$I
=
M_{r}\odot I_{r}
+
\left(1-M_{r}\right)\odot B,$
where $M_{r}$ is the rendered foreground alpha mask and
$\odot$ denotes element-wise multiplication.

\begin{figure*}[t]
    \centering
    \includegraphics[width=1.0\linewidth]{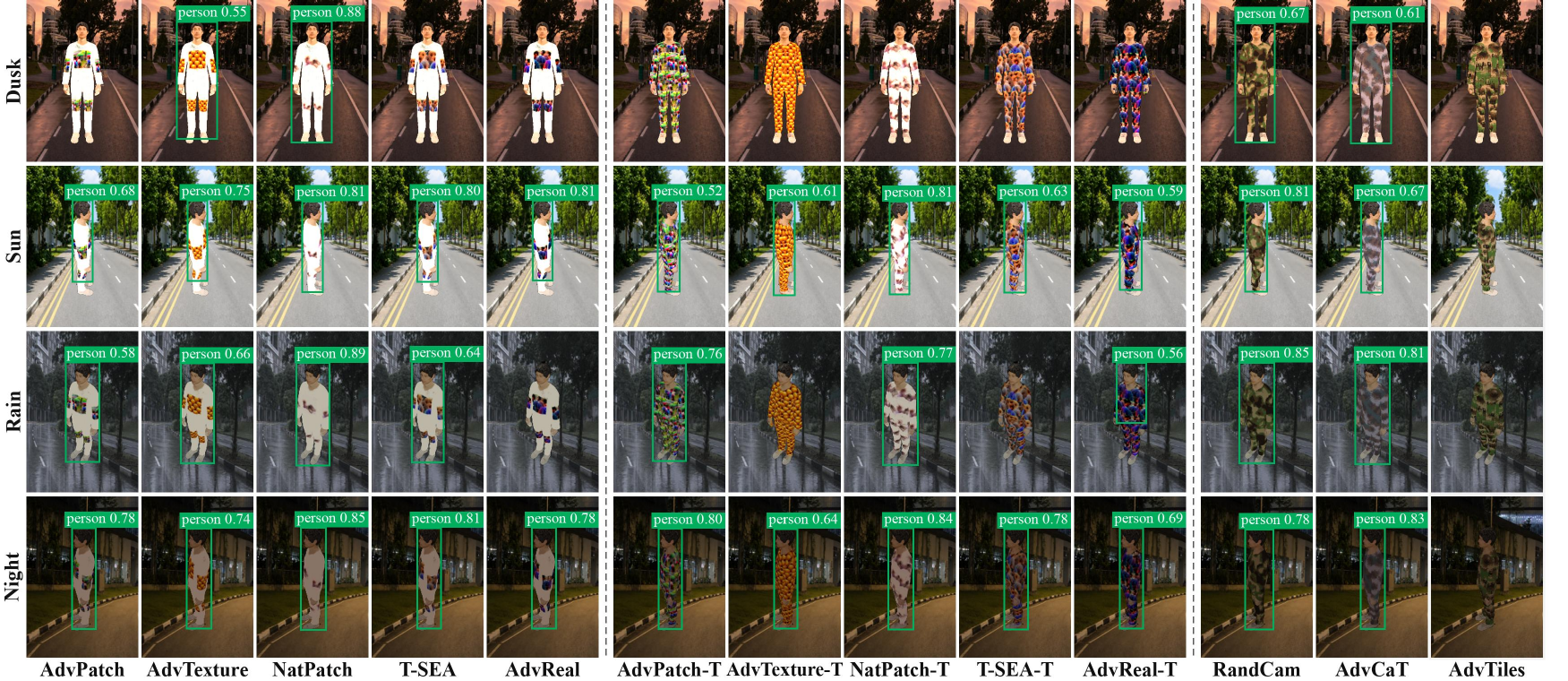}
    \vspace{-0.2in}
    \caption{Visualization of the proposed AdvTiles and other adversarial methods in the digital world.}
    \label{fig-contrast}
\end{figure*}

\subsection{Adversarial Losses}

\noindent\textbf{Detection loss.} 
To evade person detection, we minimize the confidence score of the person-class bounding box that has the highest Intersection over Union (IoU) with the ground-truth box. 
Given the synthesized image $i$, the victim detector $\mathcal{D}$ outputs a set of bounding boxes $b_q^{(i)}$, each associated with a person-class confidence score $\operatorname{Conf}_q^{(i)}$. The detection loss is defined as
\begin{equation}
\mathcal{L}_{\mathrm{det}}
=
\sum_{i}
\operatorname{Conf}_{q^*}^{(i)},
\quad
q^*
=
\operatorname*{arg\,max}_{q}
\operatorname{IoU}
\left(
gt^{(i)},
b_q^{(i)}
\right),
\label{eq:detection_loss}
\end{equation}
where $gt^{(i)}$ denotes the ground-truth bounding box of the foreground person in the synthesized image $i$.

\noindent\textbf{Diversity Loss.}  The diversity loss promotes diverse tile usage and avoids repeatedly
selecting only a few tiles. Let $u_i$ denote the average usage probability of tile $i$ computed
from the soft tile assignments.
We define the loss as
\begin{equation}
\mathcal{L}_{\mathrm{div}}
=
1-
\frac{
-\sum_{i=1}^{N}u_i\log(u_i+\epsilon)
}{
\log N
},
\end{equation}
where $N$ is the number of candidate tiles and $\epsilon$ is a small
constant for numerical stability.

\noindent\textbf{Edge Compatibility Loss.}
The edge loss encourages adjacent tiles to form compatible boundaries.
For each neighboring cell pair, we compute an expected compatibility
score by weighting the predefined compatibility of all candidate tile
pairs with their soft assignment probabilities $P$. A higher score indicates better boundary matching.
Let $y$ and $x$ denote the row and column indices of the tile grid,
respectively.
We define $s^v_{y,x}$ as the compatibility score between grid cells
$(y,x)$ and $(y+1,x)$, and $s^h_{y,x}$ as that between
$(y,x)$ and $(y,x+1)$.
Each score considers all possible tile pairs and weights their
compatibility by the corresponding assignment probabilities. Thus, we compute the edge loss:
{\small
\begin{equation}
\mathcal{L}_{\mathrm{edge}}
=
1-
\frac{1}{2}
\left(
\frac{1}{|\mathcal{N}^{v}|}
\sum_{(y,x)\in\mathcal{N}^{v}} s^{v}_{y,x}
+
\frac{1}{|\mathcal{N}^{h}|}
\sum_{(y,x)\in\mathcal{N}^{h}} s^{h}_{y,x}
\right),
\end{equation}}
where $\mathcal{N}^{v}$ and $\mathcal{N}^{h}$ denote the valid starting
positions of vertical and horizontal neighboring cell pairs,
respectively.
Minimizing this loss encourages adjacent cells to select tiles with
compatible edge patterns.

\noindent\textbf{Seam Loss.}
Although the edge loss encourages neighboring tiles to have matching
boundary patterns, it does not directly ensure color continuity at their boundaries. To improve color continuity between adjacent pixels across tile boundaries, we define the seam loss as:
\begin{equation}
\mathcal{L}_{\mathrm{seam}}
=
\frac{1}{|\Gamma|}
\sum_{\left((u,v),(u',v')\right)\in\Gamma}
\left\|
T_{:,u,v}-T_{:,u',v'}
\right\|_{1},
\end{equation}
where $T_{:,u,v}$ and $T_{:,u',v'}$ denote the RGB vectors of adjacent
pixels across a tile boundary, and $\Gamma$ denotes the set of all such
pixel pairs.

The final optimization objective is formulated as:
\begin{equation}
\mathcal{L}
=
\lambda_{\mathrm{det}}\mathcal{L}_{\mathrm{det}}
+
\lambda_{\mathrm{edge}}\mathcal{L}_{\mathrm{edge}}
+
\lambda_{\mathrm{div}}\mathcal{L}_{\mathrm{div}}
+
\lambda_{\mathrm{seam}}\mathcal{L}_{\mathrm{seam}},
\end{equation}
where $\lambda_{\mathrm{det}}$, $\lambda_{\mathrm{edge}}$,
$\lambda_{\mathrm{div}}$, and $\lambda_{\mathrm{seam}}$ denote nonnegative weights for their corresponding loss terms.

\section{Experiments}

\subsection{Experimental Setup}
\noindent\textbf{Dataset.} 
To evaluate the robustness of adversarial camouflage under diverse environmental conditions, we randomly sample 912 background images from nuScenes dataset~\cite{caesar2020nuscenes}. Each background image underwent two weather transformations (rainy and sunny) and two illumination conditions (nighttime and dusk) using IC-Light~\cite{zhang2025scaling}. We then composite the 3D human model at different viewpoints and scales onto these background images, yielding 3648 images, of which 2432 are used for training and 1216 for testing.

\noindent\textbf{Victim detectors.} 
We evaluate AdvTiles on three mainstream paradigms:
The one-stage detectors include YOLOv5~\cite{yolov5}, YOLOv8~\cite{yolov8}, YOLOv13~\cite{lei2025yolov13realtimeobjectdetection}, YOLO26~\cite{jocher2026ultralyticsyolo26unifiedrealtime}, RetinaNet~\cite{lin2017focal}, FCOS~\cite{tian2019fcos}, and SSD~\cite{liu2016ssd}. The two-stage detector is represented by Faster R-CNN~\cite{ren2015faster}. The transformer-based detectors include Deformable-DETR~\cite{zhu2020deformable} and Conditional-DETR~\cite{meng2021conditional}.

\noindent\textbf{Baselines.}
We compare AdvTiles with two kinds of representative physical adversarial attacks: 1) Patch-based attacks, which attach localized patterns to clothing, including AdvPatch~\cite{thys2019fooling}, NatPatch~\cite{hu2021naturalistic}, T-SEA~\cite{huang2023t}, AdvReal~\cite{huang2026advreal}; 2) Texture-based attacks, which optimize adversarial patterns over larger clothing region, including AdvTexture~\cite{hu2022adversarial} and AdvCaT~\cite{hu2023physically}. We also include RandCam, a randomly generated camouflage without any optimization, as a non-adversarial baseline.

\noindent\textbf{Experimental Details.}
Following prior work~\cite{huang2026advreal}, we adopt four metrics to evaluate adversarial effectiveness: Attack Success Rate (ASR)~\cite{thys2019fooling}, precision, recall, and F1-score.
We implemented AdvTiles using PyTorch. The textures are optimized for 800 epochs with a batch size of 32 using the AdamW optimizer. 
The tile generator utilized 76 valid tiles with a spatial size of $64 \times 64$ pixels. During rendering, the scale factor for clothing Gaussians is 0.3 and composited images were resized to $416 \times 416$ pixels. The loss balancing weights are set to $\lambda_{\mathrm{det}}=1.0$, $\lambda_{\mathrm{edge}}=0.5$, $\lambda_{\mathrm{div}}=0.5$, and $\lambda_{\mathrm{seam}}=0.3$.

\begin{table}[t!]
\centering
\footnotesize
\tabcolsep=1mm
\begin{tabular}{@{}lcccc@{}}
\toprule
\textbf{Method} & \textbf{ASR $\uparrow$} & \textbf{Precision $\downarrow$} & \textbf{Recall $\downarrow$} & \textbf{F1-Score $\downarrow$} \\
\midrule
AdvPatch & 12.6 ± 1.2 & 98.0 ± 0.5 & 87.4 ± 1.2 & 92.4 ± 0.9 \\
NatPatch & 37.7 ± 0.7 & 88.5 ± 1.1 & 62.3 ± 0.7 & 73.1 ± 0.4 \\
AdvTexture & 39.1 ± 1.8 & 97.1 ± 0.6 & 60.9 ± 1.8 & 74.9 ± 1.2 \\
T-SEA & 41.2 ± 2.7 & 80.6 ± 0.9 & 58.8 ± 2.7 & 68.0 ± 2.1 \\
AdvReal & \underline{72.5 ± 0.8} & \underline{24.3 ± 0.5} & \underline{27.5 ± 0.8} & \underline{25.8 ± 0.6} \\
\midrule
AdvPatch-T & 62.1 ± 1.0 & 95.6 ± 1.0 & 37.9 ± 1.0 & 54.3 ± 1.2 \\
NatPatch-T & 66.7 ± 0.7 & 95.6 ± 0.1 & 33.3 ± 0.7 & 49.4 ± 0.7 \\
AdvTexture-T & 96.9 ± 1.3 & 63.4 ± 4.0 & 3.1 ± 1.3 & 5.9 ± 2.3 \\
T-SEA-T & 94.6 ± 1.4 & 30.7 ± 2.9 & 5.4 ± 1.4 & 9.2 ± 2.2 \\
AdvReal-T & \underline{97.2 ± 0.7} & \textbf{8.4 ± 2.0} & \underline{2.8 ± 0.7} & \textbf{4.2 ± 1.0} \\
\midrule
RandCam & 12.8 ± 2.4 & 97.8 ± 0.2 & 87.2 ± 2.4 & 92.2 ± 1.3 \\
AdvCaT & 73.6 ± 0.9 & 93.3 ± 0.4 & 26.4 ± 0.9 & 41.2 ± 1.1 \\
\rowcolor{gray!15}
\textbf{AdvTiles} & \textbf{97.5 ± 0.3} & \underline{24.8 ± 10.9} & \textbf{2.5 ± 0.3} & \underline{4.5 ± 0.6} \\
\bottomrule
\end{tabular}
\vspace{-0.08in}
\caption{Performance comparison of representative adversarial attacks. The suffix ``-T'' denotes a tiled variant, in which the adversarial pattern generated by the original method is periodically repeated to cover the whole clothing.}
\label{tab-contrast}
\end{table}

\subsection{Performance in Digital and Physical Domains}
\noindent\textbf{Comparison with Baselines.} 
Table~\ref{tab-contrast}  quantitatively compares AdvTiles with representative physical adversarial attack methods against YOLOv5 under digital evaluation. As can be seen, AdvTiles achieves the SOTA performance compared to AdvPatch, AdvTexture, NatPatch, T-SEA, AdvReal and AdvCaT in terms of ASR, precision, recall and F1-Score. For example, AdvTiles achieves an ASR of 97.5\%, with a recall of 2.5\% and an F1-score of 4.5\%, substantially outperforming AdvCaT, which achieves 73.6\%, 26.4\%, and 41.2\% on the three metrics,
respectively. We observe that AdvReal-T yields a lower precision, mainly because it
induces more false-positive person detections. In contrast, AdvTiles
primarily suppresses person detections, resulting in more false negatives
and thus a substantially lower recall.

Besides the quantitative comparison, Figure~\ref{fig-contrast} presents visual comparisons between AdvTiles
and representative adversarial attacks in the digital world. It demonstrates that AdvTiles maintains more consistent attack effectiveness across diverse backgrounds and viewing angles. For example, Figure 4 shows that AdvTiles consistently suppresses person detections across different viewpoints and environmental conditions, including sunny, rainy, dusk, and nighttime scenes.

\begin{figure}[t]
    \centering
    \includegraphics[width=0.95\linewidth]{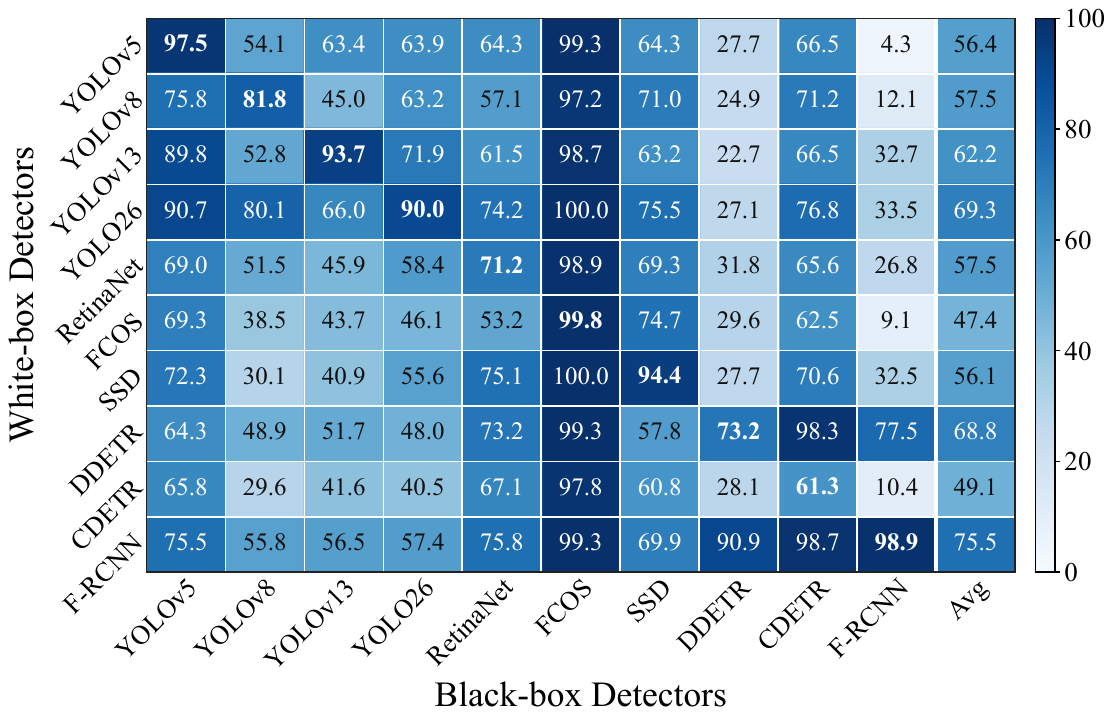}
    \vspace{-0.10in}
    \caption{Cross-detector transferability matrix of AdvTiles. Diagonal entries denote white-box attacks, while off-diagonal values indicate black-box transfers across detectors.}
    \label{fig-Confusionmatrix}
\end{figure}

\begin{figure}[t]
    \centering
    \includegraphics[width=0.99\linewidth]{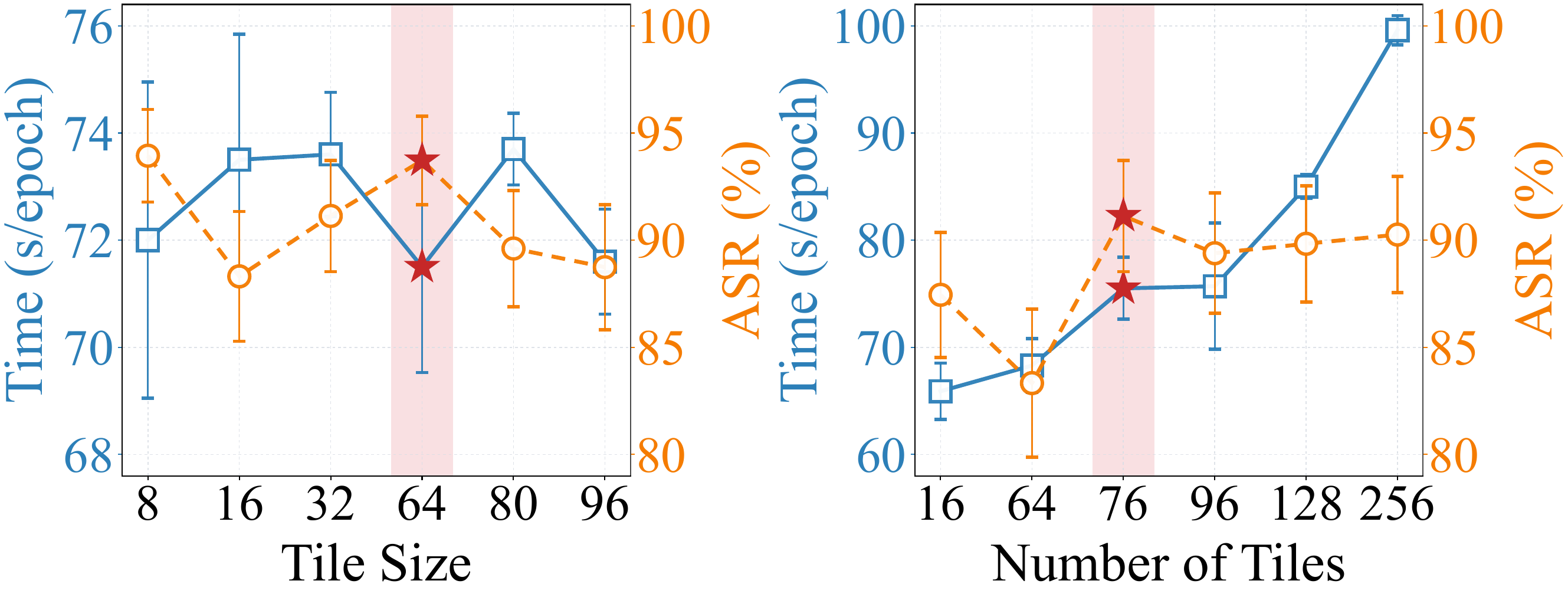}
    \vspace{-0.08in}
    \caption{Impact of tile size and number of tiles on ASR and cost.}
    \label{fig-hs}
\end{figure}

\begin{figure*}[htbp]
    \centering
    \includegraphics[width=0.95\linewidth]{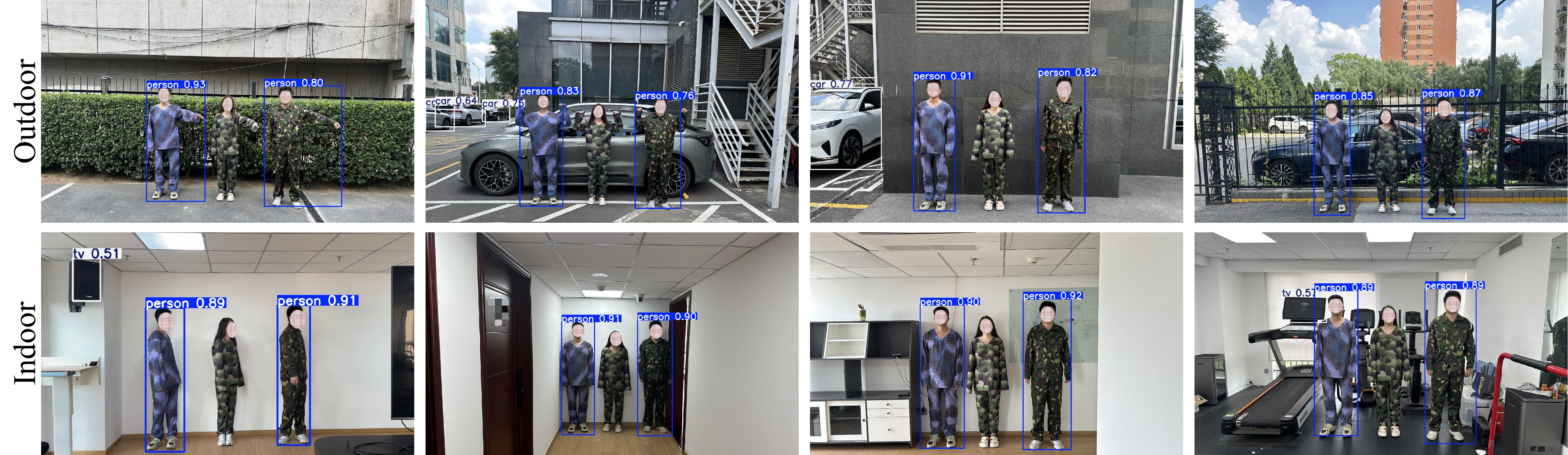}
    \vspace{-0.08in}
    \caption{Physical world comparison of AdvTiles with other camouflage clothing in indoor and outdoor environments. From left to right: AdvCaT, AdvTiles, and ordinary camouflage.
    Refer to Supplementary Materials for the video demo.}
    \label{fig-phyvis}
\end{figure*}

\begin{table*}[t]
\centering
\footnotesize
\begin{tabular}{@{}lccccc!{\vrule width 0.6pt}cccccc!{\vrule width 0.6pt}cccc@{}}
\toprule
\multirow{2}{*}{\textbf{Method}}
& \multicolumn{5}{c!{\vrule width 0.6pt}}{\textbf{Distance (m)}}
& \multicolumn{6}{c!{\vrule width 0.6pt}}{\textbf{Horizontal Viewing Angle}}
& \multicolumn{4}{c}{\textbf{Vertical Viewing Angle}} \\
\cmidrule(lr){2-6}\cmidrule(lr){7-12}\cmidrule(lr){13-16}
& \textbf{3.0} & \textbf{4.5} & \textbf{6.0} & \textbf{7.5} & \textbf{9.0}
& $\mathbf{0^\circ}$ & $\mathbf{45^\circ}$ & $\mathbf{90^\circ}$ & $\mathbf{135^\circ}$ & $\mathbf{180^\circ}$ & $\mathbf{270^\circ}$
& $\mathbf{5^\circ}$ & $\mathbf{15^\circ}$ & $\mathbf{25^\circ}$ & $\mathbf{30^\circ}$ \\
\midrule
AdvPatch & 1.4 & 6.3 & 13.9 & 21.7 & 28.1 & 6.1 & 2.9 & 26.0 & 11.8 & 16.8 & 35.2 & 6.8 & 5.8 & 5.1 & 5.1 \\
NatPatch & 41.5 & 32.3 & 31.2 & 33.4 & 36.6 & 22.9 & 19.2 & 35.7 & 33.9 & 40.9 & 42.0 & 25.1 & 22.2 & 18.1 & 16.6 \\
AdvTexture & 44.4 & 45.7 & 41.4 & 39.5 & \underline{39.5} & 48.9 & 39.1 & 45.6 & 49.8 & 52.4 & 46.0 & 54.1 & 49.0 & 36.2 & 30.4 \\
T-SEA & 66.9 & 44.4 & 35.7 & 33.7 & 34.8 & 52.9 & 33.1 & 34.7 & 38.8 & 61.2 & 36.5 & 56.0 & 51.3 & 40.5 & 33.5 \\
AdvReal & \underline{82.7} & \underline{72.7} & \underline{54.9} & \underline{43.3} & \underline{39.5} & \underline{80.8} & \underline{51.1} & 46.6 & \underline{66.5} & \underline{90.7} & 49.7 & \underline{85.5} & \underline{79.5} & \underline{58.8} & \underline{49.5} \\
\midrule
RandCam & 16.8 & 24.0 & 30.1 & 30.1 & 32.2 & 24.0 & 14.1 & 49.3 & 29.5 & 37.6 & 48.4 & 24.3 & 22.7 & 17.8 & 16.4 \\
AdvCaT & 65.2 & 40.8 & 42.0 & 33.5 & 29.1 & 49.0 & 19.6 & \underline{77.0} & 37.2 & 44.9 & \underline{66.5} & 47.0 & 20.5 & 15.2 & 15.1 \\
\rowcolor{gray!15}
\textbf{AdvTiles}
& \textbf{95.9} & \textbf{94.4} & \textbf{93.0} & \textbf{71.8} & \textbf{60.8}
& \textbf{98.3} & \textbf{97.3} & \textbf{96.1} & \textbf{95.2} & \textbf{99.7} & \textbf{98.6}
& \textbf{98.8} & \textbf{98.1} & \textbf{93.8} & \textbf{88.9} \\
\bottomrule
\end{tabular}
\vspace{-0.1in}
\caption{Performance comparison (ASR \%) across various distances, horizontal and vertical angles in the digital world.}
\label{tab-dis-angle}
\end{table*}

\begin{figure}[t]
    \centering
    \includegraphics[width=0.95\linewidth]{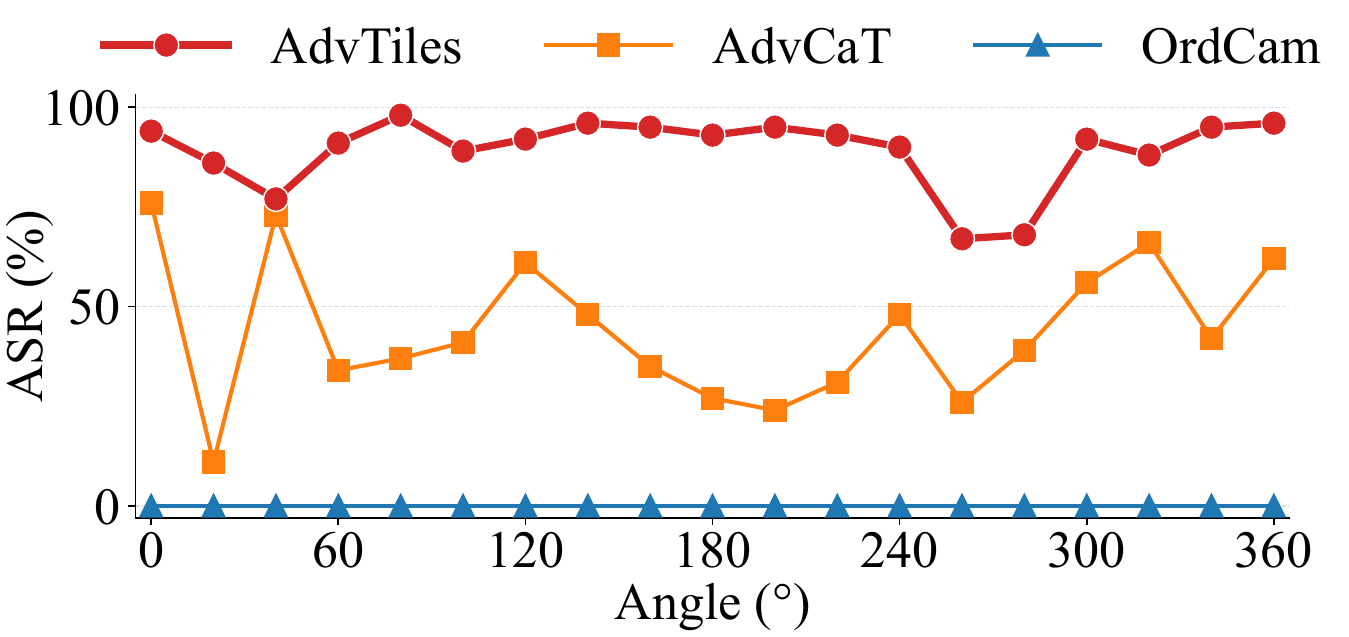}
    \vspace{-0.1in}
    \caption{Adversarial clothing across different viewing angles in the physical world. ``OrdCam'' denotes ordinary camouflage.}
    \label{fig-phycircle}
    \vspace{-0.15in}
\end{figure}

\noindent\textbf{Black-box Transferability.} 
Figure~\ref{fig-Confusionmatrix} presents the black-box transferability of AdvTiles across different detectors including YOLOv5, YOLOv8, YOLOv13, YOLOv26, RetinaNet, FCOS, SSD, Deformable-DETR, Conditional-DETR, and Faster R-CNN. Diagonal entries denote white-box attacks, while off-diagonal values indicate black-box transfers across detectors. As shown, AdvTiles still maintains attack effectiveness across black-box models. For example, the adversarial textures generated for YOLOv8 perform relatively well when transferred to other models such as YOLOv5 (75.8\%), YOLOv26 (63.2\%) and FCOS (97.2\%),  indicating strong cross-model transferability of our adversarial texture.

\noindent\textbf{Effect of Tile Size and Number.}
To evaluate the effects of tile size and tile number on attack performance
and computational efficiency, we conduct hyperparameter studies using ASR and training time, as shown in Figure~\ref{fig-hs}. For tile size, the training time remains relatively stable, while the
ASR changes significantly across different settings. A tile size of $64\times64$ achieves the highest ASR of approximately 93.8\% with a training time of 71.5 s/epoch. We therefore adopt $64\times64$ as the default tile size. For the number of tiles, the ASR varies with the tile size and reaches its peak at 76 tiles, with an ASR of approximately 91.5\%. Increasing the number of tiles beyond 76 brings no further improvement in attack performance, while the training time gradually increases. Therefore, we select 76 tiles as the default setting, providing a favorable balance between attack performance and computational efficiency.

\noindent\textbf{Robustness to Distance and Viewing Angle.} Table~\ref{tab-dis-angle} compares representative attack methods across
different distances, horizontal viewing angles, and vertical viewing angles.
Here, distance refers to the camera-to-subject distance, ranging from 3.0 to 9.0 m. Horizontal and vertical angles describe the camera viewpoint changes in the horizontal and vertical directions, ranging from $0^\circ$ to
$270^\circ$ and from $0^\circ$ to $30^\circ$, respectively. Based on the experimental results, AdvTiles maintains effective attack performance across increasing
distances and varying viewing angles. For distance, the ASR decreases
from 95.9\% at 3.0 m to 60.8\% at 9.0 m, while remaining substantially
higher than competing methods. For horizontal viewing angles, AdvTiles
maintains consistently high ASRs of over 95\% across $0^\circ$ to $270^\circ$.

\noindent\textbf{Physical Domain Validation.} Figure~\ref{fig-phyvis} presents the real-world evaluation of different camouflage clothing under indoor and outdoor scenarios. From left to right, the pedestrians wear AdvTiles-generated adversarial camouflage, AdvCat-generated adversarial camouflage, and ordinary camouflage, respectively. Among them, AdvTiles consistently achieves effective person detection evasion across
both scenarios, demonstrating the effectiveness of jointly
optimizing diverse learnable tiles for physical adversarial camouflage.

Figure~\ref{fig-phycircle} quantitatively compares the ASR of different
camouflage methods across varying viewing angles in the physical world. AdvTiles maintains consistently high attack performance over the full
$360^\circ$ range, achieving an ASR above 90\% at most viewing angles. In contrast, AdvCaT exhibits
large performance fluctuations across viewing angles, while ordinary camouflage shows
almost no attack effectiveness. These results demonstrate the strong viewpoint robustness of AdvTiles in physical environments. \textit{More physical validation is provided in the Supplementary Materials. }

\section{Conclusion} 
In this paper, we propose AdvTiles, a physical adversarial camouflage clothing against person detectors based on learnable tiles. It achieves strong attack performance while preserving a natural camouflage appearance. By introducing  Straight-through Gumbel-Softmax estimator, we enable joint optimization of tile patterns and spatial layout, providing fine-grained control over adversarial texture generation. Moreover, we leverage 3D Gaussian Splatting to improve the robustness of the adversarial camouflage across diverse viewpoints, distances, and backgrounds. Extensive digital and physical experiments demonstrate that AdvTiles
achieves strong attack performance and transferability across various detectors.  This work advances adversarial camouflage toward more natural and practical designs, taking a step toward the practical deployment of adversarial attacks in the real world.

\bibliography{aaai2027}

\end{document}